\documentclass[10pt,twocolumn,letterpaper]{article}

\usepackage{cvpr}              
\usepackage{graphicx}
\usepackage{multirow}
\usepackage{booktabs}
\usepackage[table]{xcolor}
\usepackage[dvipsnames]{xcolor}
\usepackage{wrapfig} 
\usepackage{bbding}
\usepackage{appendix}
\usepackage{xspace}
\usepackage{tocloft}
\usepackage{minitoc}
\usepackage{titletoc}
\usepackage{makecell}
\usepackage{longtable}
\usepackage{amssymb}
\usepackage{pifont}
\usepackage[accsupp]{axessibility}

\definecolor{cvprblue}{rgb}{0.21,0.49,0.74}
\usepackage[pagebackref,breaklinks,colorlinks,allcolors=cvprblue]{hyperref}

\newcommand{\smallsec}[1]{\noindent\textbf{#1}~}
\newcommand{\method}{\textsc{BOB}\xspace}
\newcommand{\sbt}[1]{\textbf{\texttt{\small #1}}}

\def\paperID{} 
\def\confName{CVPR}
\def\confYear{2026}

\title{Beyond Objects: Contextual Synthetic Data Generation for Fine-Grained Classification}

\author{
    William Yang$^{1}$ \quad Xindi Wu$^{1}$ \quad Zhiwei Deng$^{2}$ \quad Esin Tureci$^{1}$ \quad Olga Russakovsky$^{1}$ \\[5pt]
    \begin{tabular}{c}
        $^{1}$Princeton University \quad $^{2}$Google DeepMind \\
        {\tt\small \{williamyang, xindiw, esin.tureci, olgarus\}@princeton.edu} \quad {\tt\small zhiweideng@google.com}
    \end{tabular}
}
\begin{document}
\maketitle
\begin{abstract}
Text-to-image (T2I) models are increasingly used for synthetic dataset generation, but generating effective synthetic training data for classification remains challenging. Fine-tuning a T2I model with a few real examples can help improve the quality of synthetic training data; however, it may also cause overfitting and reduce diversity in the generated samples. We propose a fine-tuning strategy \method (\underline{B}eyond \underline{OB}jects) to mitigate these concerns for fine-grained classification. Given a small set of real examples, we first extract class-agnostic attributes such as scene background and object pose. We then explicitly condition on these attributes during fine-tuning of the T2I model and marginalize them out during generation. This design mitigates overfitting,  preserves the T2I model’s generative prior, reduces estimation errors, and further minimizes unintended inter-class associations. Extensive experiments across multiple T2I models, backbones, and datasets show that our method achieves state-of-the-art performance in low-shot fine-grained classification when augmented with synthetic data. Concretely, \method outperforms DataDream by 7.4\% on the Aircraft dataset (from 50.0\% to 57.4\% when fine-tuning a CLIP classifier with 5 real images augmented with 100 synthetic images). In three of the four benchmarks, fine-tuning downstream models with 5 real images augmented with \method achieves better performance than fine-tuning with 10 real images. Collectively, \method outperforms prior art in 18 of 24 experimental settings, with over 2\% accuracy improvements in 14 of these settings.
\end{abstract}    
\section{Introduction}
\label{sec:intro}
Powerful text-to-image (T2I) models trained on internet-scale datasets \citep{schuhmann2022laion, rombach2022high} have shown promise in the creation of synthetic data for representation learning \citep{tian2023stablerep, tian2024learning}, 3D synthesis \citep{poole2023dreamfusion}, and image editing \citep{hertz2022prompt, mokady2022null, brooks2023instructpix2pix}. However, there remains a considerable performance gap when using synthetic data as training data for downstream tasks such as classification \citep{burg2023image, fan2024scaling, geng2024unmet}, particularly for fine-grained classification where subtle visual differences are critical.

\begin{figure}
    \centering
    \includegraphics[width=\linewidth]{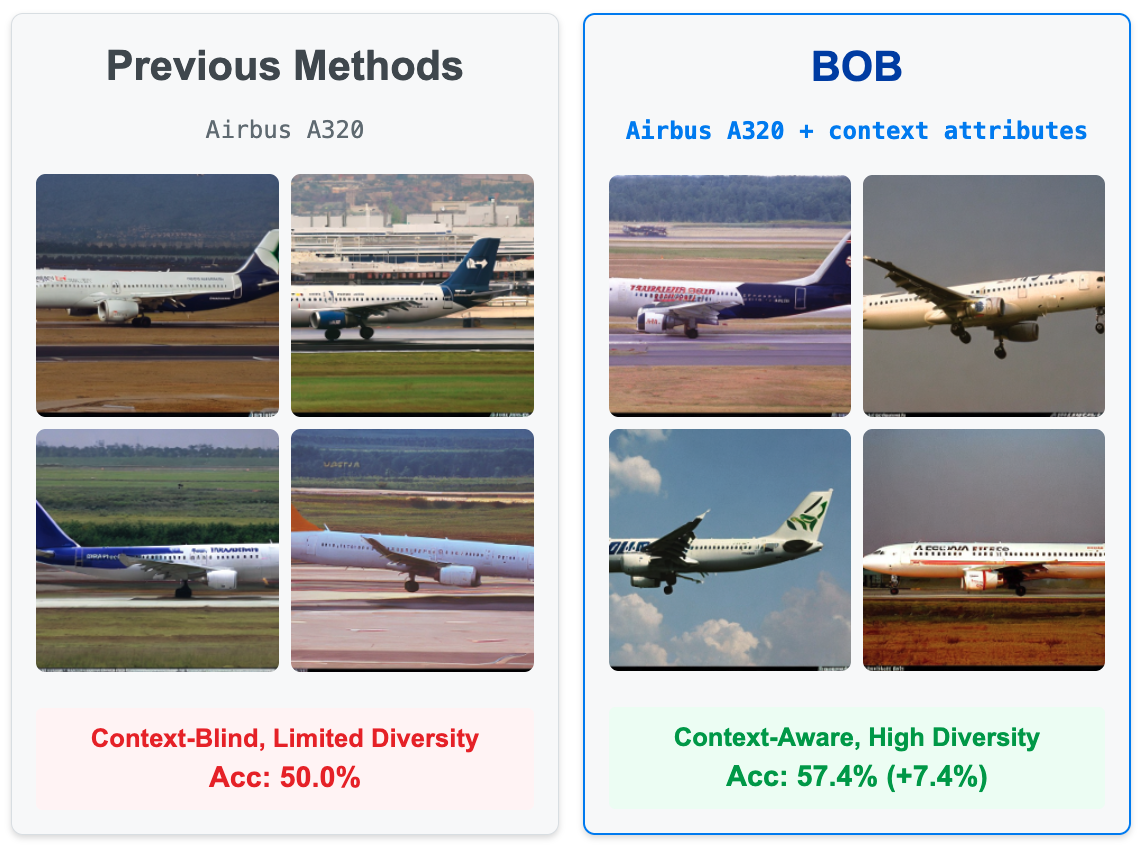}
    \vspace{-2em}
    \caption{\textbf{\method vs. prior methods.} (\textit{Left}) Prior methods do not include rich contexts during fine-tuning and sampling, leading to reduced diversity in the synthetic dataset and lower classification performance when used for downstream training. (\textit{Right}) \method mitigates this issue by incorporating rich contexts in both stages.}
    \vspace{-1em}
\end{figure}

Ideally, given a target classification task described in language (e.g., ``train an aircraft classifier to distinguish a 747-300 from a 747-400''), a T2I model can be directly used to generate training images of the desired classes. A key challenge limiting T2I models from generating informative images is the \textit{model estimation errors} caused by a misalignment between the T2I model's learned distribution and the target task \citep{geng2024unmet}. For example, the visual difference between 747-300 and 747-400 is subtle: the presence of a winglet on a slightly longer wing.
In consequence, the presence of low level artifacts and incorrect visual compositions introduce a challenge for the task of fine-grained recognition. 

One approach to mitigate model estimation errors for these classes is to provide a few real images to fine-tune the T2I model \citep{wang2024enhance,kim2024datadream}. The underlying T2I model needs to rely heavily on additional guidance from provided examples to generate not only accurate but also diverse samples to augment the training of downstream classifiers.
However, operating in the few-shot regime requires special considerations \citep{yue2020interventional}. 
The increased expressivity from fine-tuning can introduce a trade-off where the T2I model starts to overfit to the few examples, losing its strong world prior and hurting the diversity of the synthetic dataset.

In this work, we tackle fine-grained classification with synthetic data generation by introducing \method (\underline{B}eyond \underline{OB}jects). \method mitigates overfitting during T2I fine-tuning by extracting class-agnostic contextual attributes (background and pose) for each sample via a captioning model and explicitly incorporating them into the conditioning signal. During data generation, background and pose pairs are sampled across the dataset, effectively marginalizing out any unintended inter-class associations. 
We provide a comprehensive evaluation across three backbones, two T2I models, four datasets, two data scales, and seven existing methods, to demonstrate the effectiveness of \method. 

We observe the most considerable gain on Aircraft, a dataset where T2I models perform poorly and fine-tuning benefits the most. Training the CLIP model \citep{radford2021learning} using 5 real images augmented with 100 synthetic images per class, \method achieves 57.4\% accuracy compared to 50.0\% with DataDream \citep{kim2024datadream}, a 7.4\% improvement.
Importantly, across three of the four datasets (Aircraft, Car, and CUB), using 5 real images augmented with \method generated images results in \emph{better} classification performance than using 10 real images: e.g., CLIP fine-tuned on CUB achieves accuracy of 75.8\% with 5 real images  augmented with \method generated images and only 74.6\% with 10 real images without augmentation. Overall, \method outperforms existing state-of-the-art methods by at least 2\% on 18 of the 24 experimental settings (backbone, dataset source, and dataset size). On the six remaining settings (on the Pets \citep{parkhi2012cats} dataset), \method offers comparable performance with state-of-the-art. To summarize, we make the following contributions: 
\begin{enumerate}
\item We introduce stronger supervision with more detailed captioning during T2I fine-tuning to mitigate model overfitting and enhance prior preservation (\cref{sec:context_preservation}).
\item We marginalize out unintended inter-class associations by randomly sampling class-agnostic features (background, pose) across the whole dataset (\cref{sec:context_marginalization}). 
\item We provide a comprehensive evaluation (\cref{sec:experiments}) across seven previous methods, two T2I models, and 24 different experimental settings and demonstrate that our method outperforms previous methods in 18 of the 24 settings with competitive performance in the rest. 
\end{enumerate}

\section{Related Work}
\label{sec:related_work}
\smallsec{Personalization.} Many personalization methods serve as inspiration for approaches aimed at synthetic data generation for classification. Personalization methods seek to guide and control T2I models beyond text descriptions, typically using a few image exemplars. These approaches have proven effective at tailoring T2I models to reproduce highly specific visual concepts \citep{galimage, ruiz2023dreambooth, kumari2023multi, ye2023ip, zhang2023adding, li2023blip, zong2024easyref, zhaodreamdistribution}. However, while they enable strong concept-level control, their objectives differ from those required for classification-oriented synthetic data. They emphasize concept fidelity over diversity, often limiting intra-class variation and inter-class separability, which are both essential for robust classifier training. Therefore, such personalization methods are insufficient for addressing the challenges of synthetic data generation for classification.

\smallsec{Synthetic data for classification.} The field of synthetic data generation for classification has initiated a significant shift, moving toward leveraging powerful T2I models. This paradigm shift contrasts sharply with traditional data augmentation methods, such as CutMix \citep{yun2019cutmix} and Mixup \citep{zhang2018mixup}. These techniques interpolate between existing data, which helps smooth the decision function but is limited in sample diversity and fidelity. In contrast, a pre-trained T2I model provides a ``world prior" that significantly enhances both sample diversity and fidelity.

Early works demonstrated the utility of T2I models for classification. Real Guidance \citep{he2022synthetic} demonstrated that utilizing these T2I models with simple class descriptions and a few reference images can improve classification performance. Da-fusion \citep{trabuccoeffective} incorporates Textual Inversion \citep{galimage} on the few reference images to generalize to unknown concepts. Subsequent research has largely focused on two parallel areas: improving fine-tuning and enhancing generation. For fine-tuning, methods like Diff-Aug \citep{wang2024enhance}, Diff-Gen \citep{wang2024enhance}, Diff-Mix \citep{wang2024enhance} and DataDream \citep{kim2024datadream} focused on adapting different components of the T2I model (U-Net and text-encoder) to the target data. On the generation side, previous works primarily focus on better prompt design \citep{sariyildiz2023fake, yu2025diversify}, prompt augmentation with image captions \citep{dunlap2023diversify, da2023diversified}, diffusion latents interpolation \citep{zhou2023training, wang2025inversion}, leveraging vision-language models \citep{michaeli2024advancing}, and hard examples generation \citep{koohpayegani2023genie, askari-hemmat2024feedbackguided, askari-hemmat2025improving}. Additionally, \citet{liu2025does} demonstrate that unrealistic synthetic attributes often have a negligible impact on fine-tuned classifier performance. Unlike previous work that focuses primarily on fine-tuning or generation, our approach maintains T2I controllability during fine-tuning and shifts the generation objective from simple variety to marginalization to effectively mitigate spurious correlations.

\smallsec{Diffusion classifier.} Using diffusion models directly for image classification has shown promising performance \citep{li2023your, clark2023texttoimage,jaini2024intriguing, li2025generative}. This naturally raises the question of treating T2I models as classifiers which would remove the need to train a separate classifier. 
However, inference with diffusion classifiers is very computationally expensive, increasing each ImageNet classification from seconds to over 10 minutes \citep{russakovsky2015imagenet, li2023your}, motivating research on distilling these capabilities into efficient downstream classifiers.
\begin{figure*}[t]
    \centering
    \includegraphics[width=0.99\linewidth]{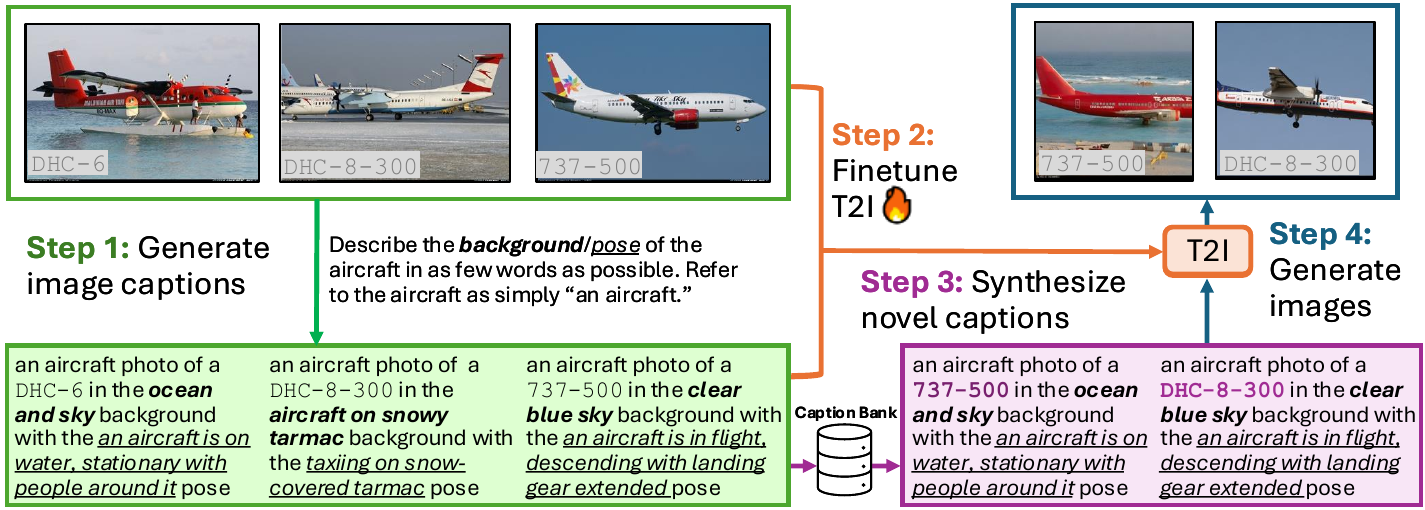}
    \caption{\textbf{Overview of \method.} We extract background and pose attributes from training images using a captioning model (\textit{Step 1)}, apply \textbf{context preservation} by fine-tuning the T2I model with enriched captions containing class names and context attributes (\textit{Step 2}), and then perform \textbf{context marginalization} by generating synthetic data through randomly sampling background-pose pairs across the entire dataset (\textit{Step 3-4}). This preserves class-relevant features while reducing spurious class-context associations.\
    }
    \label{fig:method}
    \vspace{-1em}
\end{figure*}

\section{Methodology}
\label{sec:method}

We propose an effective approach for fine-grained synthetic data generation that addresses the critical problem of overfitting during T2I model fine-tuning and image generation (\cref{fig:method}). This overfitting exists in both modalities: in the texts, the classification task's ambiguous text-to-image mapping inherently loses T2I controllability by failing to articulate the intra-class visual range; and in the images, the few-shot regime fosters unintended inter-class associations. Our method mitigates both overfitting issues in two stages: \emph{Context Preservation} during fine-tuning and \emph{Context Marginalization} during data generation. We describe both in detail in this section.

\subsection{Context Preservation} 
\label{sec:context_preservation}
Our first mitigation strategy \emph{Context Preservation} aims to recover the intra-class visual range lost in the text modality. While T2I models require detailed text-to-image mappings, classification datasets only provide generic labels. Previous approaches use class-specific templates (e.g., ``a photo of a \sbt{[classname]}") but this simplification reduces diverse visual information to a single description, leading to a loss of T2I controllability. To address this, we propose associating each image with a unique, descriptive text. We achieve this by extracting and explicitly encoding class-agnostic attributes (background and pose) into the text conditioning. This enables the model to learn the specific association between these attributes and the visual context during fine-tuning. Concretely, each image is associated with a unique caption following the template:

\begin{center}
\textbf{a \sbt{[descriptor]} photo of a \sbt{[classname]} in the \sbt{[background]} background with the \sbt{[pose]} pose}.     
\end{center}

The \sbt{[descriptor]} is a dataset-level general descriptor such as aircraft or birds. \sbt{[classname]} is the name of the class provided by the dataset. 
The background/pose is extracted by a captioning model for each image. We leverage the Qwen 2.5VL-7B, a state-of-the-art vision-language model \citep{Qwen2.5-VL} to extract the background with the following prompt: \textbf{describe the background of the [descriptor] in as few words as possible. Refer to the [descriptor] as simply `a [descriptor]'}. 
Similarly, we use the same prompt with `background' replaced by `pose' to extract the pose. 

We also store the extracted background and pose into a caption bank $\mathcal{B} = \{(b_i, p_i)\}_{i=1}^N$, where $b_i$ and $p_i$ represent the background and pose attributes of the $i$-th training image, respectively (\cref{fig:method}, steps 1-2). 
This prompting approach serves two purposes: (1) it provides necessary context to guide accurate attribute extraction, and (2) it prevents potential leakage of class-specific information by maintaining generic references to the object category.

Once we have established the image-text pairing, we fine-tune the diffusion model using the standard diffusion objective.
We follow the standard parameter-efficient fine-tuning procedure by using Low-Rank Adaptation (LoRA) \citep{hu2022lora} to fine-tune the attention layers of both the U-Net \citep{ronneberger2015u} and CLIP text encoder \citep{radford2021learning}.
Consider the following notations: $\theta$ as parameters of the attention layers, the image as $x$, the text as $y$, the CLIP text encoder as $c(y)$, timestep of diffusion process as $t$, and the U-Net model  as $\epsilon_\theta(x, c_\theta (y), t)$. The parameters $\theta$ are updated by minimizing the following objective: $
 \mathbb{E}_{(x,y) \sim D,\, \epsilon \sim \mathcal{N},\, t \sim \mathcal{U}} \| \epsilon - \epsilon_\theta(x, c_\theta(y), t) \|_2^2.
$

\subsection{Context Marginalization} 
\label{sec:context_marginalization}

Our second stage \emph{Context Marginalization} reduces unintended associations between class labels and visual context which are common in data-scarce settings. When only a small number of images are available per class, certain backgrounds or poses may appear only with specific labels due to limited coverage. The central intuition behind Context Marginalization is that the contextual attributes retained during fine-tuning can be reused during generation to prevent the model from reinforcing the dataset idiosyncrasies.

\setlength{\columnsep}{10pt}
\begin{wrapfigure}{r}{0.35\linewidth}
    \centering
    \vspace{-1em}
    \includegraphics[width=\linewidth]{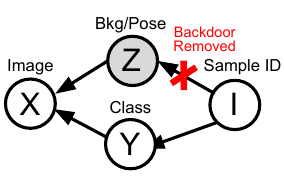}
    \caption{Causal graph}
    \label{fig:graph}
    \vspace{-1em}
\end{wrapfigure} 
We can formalize this problem more rigorously from a causal perspective. Consider the generation process of our training data that our T2I model emulates shown in Fig. \ref{fig:graph}: image $X$ is generated given the class-relevant attributes $Y$ and the class-agnostic attributes $Z$. 
To directly model the relationship between $X$ and $Y$, 
we sample from the intervention distribution $P(X|do(Y))$ by invoking the back-door adjustment: $P(X|do(Y)) = \sum_Z P(X|Y,Z)P(Z).$

To sample from $P(Z)$, we generate synthetic images using the same prompt template but randomly sample background–pose pairs $(b,p)$ from our caption bank $\mathcal{B}$, independent of the class label (steps 3-4 in \cref{fig:method}). This sampling strategy exposes each class to the variety of backgrounds and poses present in the dataset, helping the model focus on class-defining visual features rather than incidental context.

\section{Experiments}
\label{sec:experiments}
In this section, we will first go over the experimental setup for our comprehensive evaluation. Next, we will perform detailed quantitative analysis on the experimental results to demonstrate advantages of our proposed approach with respect to previous methods. 
Lastly, we present additional analysis showing that our method produces synthetic data that better aligns with the target data distribution and perform ablations showing the necessity of both the context preservation and content marginalization for generating informative data for the downstream classification task.

\subsection{Experimental Setup}
\label{sec:experimental_setup}
\smallsec{Datasets.} We follow the standard settings for data-scarce augmentation where we only have 5 or 10 real images per class. We follow the previous evaluation setting \citep{wang2025inversion} where we use Aircraft \citep{maji13fine-grained}, CUB \citep{wah2011caltech}, Car \citep{krause20133d}, and Pets \citep{parkhi2012cats} datasets for evaluation. Additionally, we also use the CUB-LT \citep{samuel2021generalized} and Flower-LT \citep{wang2024enhance} datasets to extend \method to a different data-scarce setting: long-tail classification.

\smallsec{Backbones.} We use three backbones with different degrees of language supervision during pre-training for downstream task fine-tuning. For a backbone with dense language supervision, we use the CLIP VIT-B/16 model \citep{dosovitskiy2020image, radford2021learning}. For a backbone with weak language supervision, we use the ImageNet classification-trained ResNet-50 model \citep{he2016deep}. Finally, for a backbone with no language supervision, we use the masked auto-encoder (MAE) VIT-B/16 model pre-trained on ImageNet \citep{he2022masked}. \emph{Unlike previous works which focus on one of these types, we evaluate across all three language supervision settings to provide a wider perspective towards the behavior and usefulness of different methods.} 

\smallsec{Baseline methods.} We use seven popular existing data generation or augmentation methods for comparisons: RealGuidance \citep{he2022synthetic}, Da-fusion \citep{trabuccoeffective}, Diff-Aug \citep{wang2024enhance}, Diff-Gen \citep{wang2024enhance}, Diff-Mix \citep{wang2024enhance}, DataDream \citep{kim2024datadream}, and Diff-II \citep{wang2025inversion}. In the synthetic data generation stage, we use the hyperparameters and procedure provided in the original paper. We utilize the fine-tuned T2I weights provided by the Diff-II paper for synthetic data generation on Stable Diffusion v1.5 \citep{rombach2022high}. We reproduce Diff-Aug, Diff-Gen, and Diff-Mix using the same T2I model from the paper: Stable Diffusion v1.5. Similarly for DataDream, we reproduce their results using the same T2I model: Stable Diffusion v2.1-base. Since RealGuidance and Da-fusion are relatively older methods on older T2I models, we reproduce their results with the more recent T2I model of Stable Diffusion v2.1-base. In summary, when possible, we utilized the relevant T2I model for each prior method to compare with directly. 

\smallsec{Implementation details.} For fair comparison with existing methods, we fine-tune our method on both Stable Diffusion v1.5 and Stable Diffusion v2.1-base. We utilize the same hyperparameters as DataDream with the exception of longer fine-tuning: 400 epochs instead of 200 epochs. For fair comparisons, we also extend the DataDream method to 400 epochs. We show in the Appendix that both methods improve with the increased training. 

The classification objective function is a weighted average on the cross-entropy loss between real ($r$) and synthetic data ($s$): $L = \lambda \cdot CE(f_\theta(x_r, y_r) + (1 - \lambda) \cdot CE(f_\theta(x_s), y_s) $,

where $\lambda$ is a hyperparameter and $f_\theta(x)$ is the classifier with parameters $\theta$. Mixup (except for the Diff-Mix setting) and Cutmix augmentation is applied separately between the real and synthetic data. For fair comparison, we perform hyperparameter tuning by training only for 10 epochs and evaluating a separate validation set before finally training the downstream classifier with the best hyperparameters for 100 epochs. Details are provided in the Appendix.

\begin{table*}[t]
\centering
\caption{\textbf{Few-shot classification accuracy.} The best performing method is in \textbf{bold} and the second best is \underline{underlined}. Across three different backbones used as downstream classifier, our method outperforms existing methods by a considerable margin on Aircraft (AirC), Car, and CUB. On the Pets dataset, our method obtains similar performance of previous methods.}
\vspace{-1em}

\resizebox{\textwidth}{!}{%

\begin{tabular}{llc *{5}{c} *{5}{c}}
\toprule
& \textbf{Method} & \multirow{2}{*}{\makecell{\textbf{SD}\\\textbf{Ver.}}} &
\multicolumn{5}{c}{\textbf{5-shot}} &
\multicolumn{5}{c}{\textbf{10-shot}} \\ 
\cmidrule(lr){4-8} \cmidrule(lr){9-13}
& & & AirC & Car & CUB & Pets & Avg &  AirC & Car & CUB & Pets & Avg \\ 
\midrule

\multirow{9}{*}{\rotatebox{90}{\textbf{CLIP}}} 
& Real Only       &      &   44.37    &   79.01   &   67.72   &  \underline{92.76}   & 70.97 & 55.73    &  84.87  &   74.59  &  93.65  & 77.21  \\
& Diff-Aug        & v1.5   &  44.67    &   80.93   &    68.05 &   92.27 & 71.48{\color{OliveGreen}(+0.51)}  &   57.19   &  86.07   &  77.29    &   93.44  & 78.50{\color{OliveGreen}(+1.29)} \\
& Diff-Gen        & v1.5   & 47.54   &   81.60   &   69.21  &   91.69 &  72.51{\color{OliveGreen}(+1.54)} &  58.60  &   88.43   &  76.45   &    93.57 & 79.26{\color{OliveGreen}(+2.05)} \\
& Diff-Mix        & v1.5   & 42.09  &  80.19   &  67.45   &   92.78  & 70.63{\color{BrickRed}(-0.34)} &  52.73   &  87.31    &  73.60  &    93.34  & 76.75{\color{BrickRed}(-0.47)}\\
& Diff-II         & v1.5   &  49.02 &  82.16    &  70.41  &  92.75  & 73.59{\color{OliveGreen}(+2.62)}  &  60.25  & 89.02  &  77.05  &   93.02 & 79.84{\color{OliveGreen}(+2.63)}  \\
& \cellcolor{green!14}\method (ours) & \cellcolor{green!14}v1.5   
& \cellcolor{green!14}\underline{55.85} & \cellcolor{green!14}\underline{88.10} & \cellcolor{green!14}\textbf{75.84} & \cellcolor{green!14}92.24 & \cellcolor{green!14} \underline{78.01{\color{OliveGreen}(+7.04)}}
& \cellcolor{green!14}\textbf{68.88}  & \cellcolor{green!14}\textbf{92.42} & \cellcolor{green!14}\textbf{81.26} & \cellcolor{green!14}93.31 & \cellcolor{green!14} \textbf{ 83.97{\color{OliveGreen}(+6.76)}}\\
& RealGuidance    & v2.1   &  43.12   &  80.23   &  69.93  &   \textbf{92.78}   & 71.52{\color{OliveGreen}(+0.55)} &  52.96    &   85.36   & 76.45     &  92.79 & 76.89{\color{BrickRed}(-0.32)}   \\
& Da-fusion       & v2.1   &  42.39   &  79.83  &  69.33   &   92.59  & 71.035{\color{OliveGreen}(+0.07)} &  55.27    &  79.83   &  76.02    &   \textbf{94.04} & 76.29{\color{BrickRed}(-0.92)}  \\
& DataDream       & v2.1   &  50.04   & 84.58  &  70.74    &  92.67 & 74.51{\color{OliveGreen}(+3.54)} &  63.89  & 90.26  &   78.90    &  93.90 & 81.74{\color{OliveGreen}(+4.53)}  \\
& \cellcolor{green!14}\method (ours) & \cellcolor{green!14}v2.1   
& \cellcolor{green!14}\textbf{57.37} & \cellcolor{green!14}\textbf{88.41} & \cellcolor{green!14}\underline{75.43} & \cellcolor{green!14}92.73 & \cellcolor{green!14} \textbf{78.49{\color{OliveGreen}(+7.52)}}
& \cellcolor{green!14}\underline{67.61} & \cellcolor{green!14}\underline{92.00} & \cellcolor{green!14}\underline{80.95} & \cellcolor{green!14}\underline{93.77} & \cellcolor{green!14} \underline{83.58{\color{OliveGreen}(+6.37)}}\\
\midrule

\multirow{9}{*}{\rotatebox{90}{\textbf{ImageNet}}} 
& Real Only       &        &  39.62    &   56.16   & 48.22  &  83.17 & 56.79    &   55.48  &   78.50    &   68.05   &    86.75 & 72.20 \\
& Diff-Aug        & v1.5   &    43.27   &  70.95  &  57.24   &   85.09 & 64.14{\color{OliveGreen}(+7.35)} &  57.91   &   85.34  &  72.74   &    87.40 & 75.85{\color{OliveGreen}(+3.65)}  \\
& Diff-Gen        & v1.5   &   48.42  &  80.73    &  60.91    &  86.95  & 69.25{\color{OliveGreen}(+12.46)}  &  60.32   &  88.85    &  72.40   &   \underline{89.93} & 77.88{\color{OliveGreen}(+5.68)}  \\
& Diff-Mix        & v1.5   &  38.27 &  76.58   &   53.28  &  85.36 & 63.37{\color{OliveGreen}(+6.58)}   &   52.21 &    86.41  &  68.16  &   88.63  & 73.85{\color{OliveGreen}(+1.65)} \\
& Diff-II         & v1.5   &  52.28   &  82.95   &  63.60    &   \textbf{87.63} & 71.62{\color{OliveGreen}(+14.83)} &  62.81   &   88.53   &  73.60    &   \textbf{89.95}  & 78.72{\color{OliveGreen}(+6.52)} \\
& \cellcolor{green!14}\method (ours) & \cellcolor{green!14}v1.5   
& \cellcolor{green!14}\underline{60.02} & \cellcolor{green!14}\textbf{88.80} & \cellcolor{green!14}\underline{68.78} & \cellcolor{green!14}86.38 & \cellcolor{green!14} \underline{76.00{\color{OliveGreen}(+19.21)}}
& \cellcolor{green!14}\underline{70.79} & \cellcolor{green!14}\underline{92.60} & \cellcolor{green!14}\underline{78.62} & \cellcolor{green!14}89.04 & \cellcolor{green!14} \underline{82.76{\color{OliveGreen}(+10.56)}} \\
& RealGuidance    & v2.1   &   35.53   &  68.76    &   57.34  &   \underline{87.25} & 62.22{\color{OliveGreen}(+5.43)}  & 49.23   &  83.13    & 70.43     &  87.23  & 72.51{\color{OliveGreen}(+0.31)}  \\
& Da-fusion       & v2.1   &    42.60 &  73.99   &  59.03    &   86.17 & 65.45{\color{OliveGreen}(+8.66)}  &   56.69   &  85.80   &   71.38   &   88.98  &75.71{\color{OliveGreen}(+3.51)} \\
& DataDream       & v2.1   &  54.58   &  86.15    &   67.40  &   84.85 & 73.25{\color{OliveGreen}(+16.46)}  &   67.99  & 91.29   & 77.48    & 88.38 & 81.29{\color{OliveGreen}(+9.09)}\\
& \cellcolor{green!14}\method (ours) & \cellcolor{green!14}v2.1   
& \cellcolor{green!14}\textbf{60.31} & \cellcolor{green!14}\underline{88.64} & \cellcolor{green!14}\textbf{71.38} & \cellcolor{green!14}87.00 & \cellcolor{green!14} \textbf{76.83{\color{OliveGreen}(+20.04)}}
& \cellcolor{green!14}\textbf{73.78} & \cellcolor{green!14}\textbf{92.52} & \cellcolor{green!14}\textbf{79.52} & \cellcolor{green!14}89.40 & \cellcolor{green!14} \textbf{83.81{\color{OliveGreen}(+11.61)}}\\
\midrule

\multirow{9}{*}{\rotatebox{90}{\textbf{MAE}}} 
& Real Only       &        &   41.13   &  53.94  &  39.63   &  76.81  & 52.88  &   57.61   &  79.12    &  62.50    &    82.97 & 70.55 \\
& Diff-Aug        & v1.5   &   44.28   &   72.22   &  55.35    &  74.76    & 61.65{\color{OliveGreen}(+8.77)} &   60.64   &  86.79    &  75.72    &  84.41  & 76.89{\color{OliveGreen}(+6.34)} \\
& Diff-Gen        & v1.5   &   51.79   &   82.66   &  62.79    &  77.32  & 68.64{\color{OliveGreen}(+15.76)}  &   63.85   &   90.92   &   77.10   &   85.15 & 79.26{\color{OliveGreen}(+8.71)}  \\
& Diff-Mix        & v1.5   & 41.46  &  78.16   &   52.50  &  81.28  & 63.35{\color{OliveGreen}(+10.47)}  &  60.31  &   88.29   &  69.09  &   86.40 & 76.02{\color{OliveGreen}(+5.47)}  \\
& Diff-II         & v1.5   &   54.90   &   82.09   &  66.53    &  \textbf{88.33}  & 72.69{\color{OliveGreen}(+20.08)} &   65.20   &   90.39  &   77.05   &    \underline{89.21} & 80.46{\color{OliveGreen}(+9.91)} \\
& \cellcolor{green!14}\method (ours) & \cellcolor{green!14}v1.5   
& \cellcolor{green!14}\textbf{62.32} & \cellcolor{green!14}\underline{87.73} & \cellcolor{green!14}\underline{69.23} & \cellcolor{green!14}\underline{87.46} & \cellcolor{green!14} \underline{76.69{\color{OliveGreen}(+23.81)}}
& \cellcolor{green!14}\underline{75.70} & \cellcolor{green!14}\textbf{93.16} & \cellcolor{green!14}\underline{80.17} & \cellcolor{green!14}\textbf{89.56} & \cellcolor{green!14} \underline{84.65{\color{OliveGreen}(+14.10)}} \\
& RealGuidance    & v2.1   &  38.70    &  68.78    &   52.47   &   80.62 & 60.14{\color{OliveGreen}(+7.26)}  &   57.13   &  84.58    &   73.33   &  86.94  & 75.50{\color{OliveGreen}(+4.95)} \\
& Da-fusion       & v2.1   &  46.98    &  73.39    &   51.90   &   75.52  & 61.95{\color{OliveGreen}(+9.07)} &   58.57   &   87.61   &   73.33   &   83.21  & 75.68{\color{OliveGreen}(+5.13)} \\
& DataDream       & v2.1   &   58.54   &  85.81    &  69.07    &    80.38   & 73.45{\color{OliveGreen}(+20.57)} & 71.20    &  92.12    &   79.15   &   86.35  & 82.21{\color{OliveGreen}(+11.66)} \\
& \cellcolor{green!14}\method (ours) & \cellcolor{green!14}v2.1   
& \cellcolor{green!14}\underline{61.21} & \cellcolor{green!14}\textbf{88.48} & \cellcolor{green!14}\textbf{73.21} & \cellcolor{green!14}86.72 & \cellcolor{green!14} \textbf{77.41{\color{OliveGreen}(+24.53)}}
& \cellcolor{green!14}\textbf{75.85} & \cellcolor{green!14}\underline{92.96} & \cellcolor{green!14}\textbf{81.29} & \cellcolor{green!14}88.99 & \cellcolor{green!14} \textbf{84.77{\color{OliveGreen}(+14.22)}} \\
\bottomrule
\end{tabular}
}
\vspace{-10pt}
\label{table:fewshot}
\end{table*}

\subsection{Few-shot Classification} 
In the few-shot classification setting we use 5 or 10 real images per class which we use to fine-tune the pre-trained T2I model before generating 100 synthetic images per class. Additional experiments using different numbers of real images and synthetic images are shown in the Appendix. In training the downstream classifier, the real images are replicated such that there is a 50/50 split between synthetic and real data.
We present the performance of our method compared with seven existing baselines in \cref{table:fewshot}: Diff-Aug, Diff-Gen, Diff-Mix, Diff-II using Stable Diffusion v1.5 and RealGuidance, Da-fusion, and DataDream using Stable Diffusion v2.1. 
Downstream tasks include Aircraft, a task with lowest maximum starting baseline performance of 44.37\%, moderate maximum baseline performance tasks of Car and CUB (79.01\% and 67.72\%), to Pets, with a relatively high maximum baseline performance of  92.76\% in the 5-shot setting.
Our method improves performance over all the baseline and the best performing existing method, in all tasks with the exception of Pets.
For Aircraft, Car and CUB downstream tasks, \method improves performance by at least 6.36\% and up to 34.54\% over the baseline of training with only the real data, and at least 2.77\% and up to 10.25\% over the best performing existing method. Detailed analysis focusing on specific aspects of these experiments follow.

\begin{table*}[]

\caption{\textbf{Long-tail classification accuracy.} The best performing method is in \textbf{bold} and the second best is \underline{underlined}. The expected accuracy across all the classes is reported. \textit{Many} reports classes with over 20 (30) examples for CUB-LT (Flower-LT). \textit{Medium} reports classes with between 5-20 (10-30) examples for CUB-LT (Flower-LT). \textit{Few} reports classes with under 5 (10) examples for CUB-LT (Flower-LT). Results from fine-tuning an ImageNet pre-trained ResNet-50 indicates that \method outperforms existing methods.}
\vspace{-1em}
\centering
\resizebox{\textwidth}{!}{
\begin{tabular}{lccccccccccccc}
\toprule
\multirow{3}{*}{Method} & \multirow{3}{*}[-0.5ex]{\centering\makecell{SD\\ Ver.}}
 & \multicolumn{6}{c}{CUB-LT} & \multicolumn{6}{c}{Flower-LT} \\ 
\cmidrule(lr){3-8}\cmidrule(lr){9-14} 
                        &                       & \multicolumn{4}{c}{\textbf{IF=100}} & \multirow{2}{*}{\textbf{50}} & \multirow{2}{*}{\textbf{10}} & \multicolumn{4}{c}{\textbf{IF=100}} & \multirow{2}{*}{\textbf{50}} & \multirow{2}{*}{\textbf{10}} \\
                        &                       & Many    & Med    & Few   & All   &                              &                              & Many    & Med    & Few   & All   &                              &                              \\ 
\midrule
Real Only               &                       &     86.00    &  65.22     &   17.84    &  37.73     &     49.32        &      60.09                        &    99.45     &     97.70      &    60.74   &   72.08    &                  87.41            &          93.70                   \\
Diff-Gen                & v1.5                   &   87.22      &    68.69       &   26.06    &  43.95     &        59.47          &         67.78                   &     99.79    &     96.71      &    71.25   &   79.17    &                 92.93             &              95.12                \\
Diff-Mix                & v1.5                   & 87.70    &     73.12    &    32.76   &   49.46   &    60.61     &     67.06      &   99.61  &     98.47     & 73.17  & 80.93   &     91.99           &        94.77               \\
Diff-II                 & v1.5                   &    87.54    &     72.16     &    44.05   &   56.10   &   64.52   &      70.28           &     99.82    &     98.45      &   79.51    &    85.35   &                 95.20             &               \underline{97.62}               \\
\cellcolor{green!14}\method (ours)          & \cellcolor{green!14}v1.5                   &    \cellcolor{green!14}\textbf{88.48}     &      \cellcolor{green!14}\underline{75.37}     &   \cellcolor{green!14}\underline{52.24}    &  \cellcolor{green!14}\underline{62.19}   &                  \cellcolor{green!14}\underline{70.57}       &         \cellcolor{green!14}\underline{74.54}                     &   \cellcolor{green!14}\textbf{100.0}      &    \cellcolor{green!14}\underline{98.56}      &    \cellcolor{green!14}\textbf{84.13}   &   \cellcolor{green!14}\textbf{88.60}    &              \cellcolor{green!14}\underline{95.68}                &          \cellcolor{green!14}96.13                    \\
DataDream               & v2.1                   &    87.25     &     71.23      &  39.72     &  53.42     &             66.05                 &   72.32                           &    \underline{100.0}     &       \textbf{98.67}    &   79.96   &  85.73     &                 94.13             &             96.08                 \\
\cellcolor{green!14}\method (ours)          & \cellcolor{green!14}v2.1                   &     \cellcolor{green!14}\underline{88.43}    &   \cellcolor{green!14}\textbf{75.56}  &    \cellcolor{green!14}\textbf{53.47}   &   \cellcolor{green!14}\textbf{63.06}     &     \cellcolor{green!14}\textbf{73.00}              &     \cellcolor{green!14}\textbf{76.28}                         &     \cellcolor{green!14}99.45    &     \cellcolor{green!14}98.41      &    \cellcolor{green!14}\underline{83.48}   &   \cellcolor{green!14}\underline{88.07}    &             \cellcolor{green!14}\textbf{96.85}                 &       \cellcolor{green!14}\textbf{97.80}                       \\ \hline
\end{tabular}
}
\label{table:longtail}
\vspace{-10pt}
\end{table*}

\smallsec{Aircraft classification task.} The pre-trained Stable Diffusion model has the least amount of knowledge about the Aircraft dataset, as indicated by the very poor performance of RealGuidance which is a personalization and fine-tuning free method. 
Focusing on the 5-shot setting for the FGVC-Aircraft classification task, using the ImageNet trained ResNet-50,
augmenting real images with RealGuidance generated images results in a degradation in performance of the ImageNet pretrained model by 4.09\% and 6.25\% in 5-shot and 10-shot settings. 
Improvements by other previous methods range in 3.65-14.96\% with DataDream performing the best, while our method, \method leads to a 20.69\% improvement raising the accuracy from 39.62\% to 60.31\%, 5.73\% higher improvement than DataDream.
Including the CLIP and MAE backbones for downstream tasks, \method provides 3.78-7.33\%  improvement in the 5-shot and 
4.65-5.79\% in the 10-shot settings over the best performing previous method for this downstream task.

\smallsec{Pets classification task.}
Pets classification task has the highest baseline performance of 76.81\%, 83.17\% and 92.76\% with 5 real images and 82.97\%, 86.75\% and 93.65\% with 10 real images for the MAE, ImageNet and CLIP backbones indicating that this downstream dataset distribution is 
represented much better compared to other datasets. We note two interesting observations. First, it appears that the Stable Diffusion model have additional knowledge of this dataset since RealGuidance can improve performance by more than 4\%. Second, there is low variability in performance across all the methods in CLIP and ImageNet backbones. This is most likely due to significant overlap from the pre-training data. We comment on this in more detail in the Appendix.
Overall, our method \method improves performance on par with existing methods in this task reaching a performance of 87.46\%, 87\% and 92.73\% in the 5-shot and 89.56\%, 89.40\% and 93.77\% in the 10-shot setting, within 1\% of performance of best existing method.

\smallsec{Comparison of 5-shot and 10-shot performance.} 
We compare the performance of training with 5 real images augmented with synthetic data generated from those 5 images against training with 10 real images alone (without synthetic augmentation).
This comparison is particularly revealing as it demonstrates whether our synthetic data generation method can effectively substitute for collecting additional real-world training examples, which is often expensive and time-consuming in practice.
As shown in \cref{table:fewshot}, we observe a consistent and striking pattern across all three backbones (CLIP, ImageNet, and MAE): the addition of synthetic data generated by fine-tuning with only 5 real images using \method outperforms using 10 real images alone in all datasets with the exception of Pets, where the performance is similar.
For the Aircraft dataset, the gains are particularly substantial. With the ImageNet backbone, 5 real images plus \method generated synthetic data (v2.1) achieves 60.31\% accuracy, substantially outperforming the 55.48\% achieved with 10 real images alone with a gain of 4.83\%. With the MAE backbone, the improvement is even more pronounced: 61.21\% (5-shot + \method) versus 57.61\% (10-shot real only), representing a 3.60\% gain.
Similarly, for the CUB dataset with the CLIP backbone, our 5-shot augmented approach reaches 75.43-75.84\% accuracy compared to 74.59\% with 10 real images, and for the Car dataset with the ImageNet backbone, we achieve 88.64\% with 5-shot + \method versus 78.50\% with 10-shot real only, which is an impressive 10.14\% improvement.
The only exception is the Pets dataset, where the 10-shot real-only baseline already achieves very high performance, leaving limited room for improvement and suggesting that the pre-trained models already have substantial knowledge of this domain.
These results indicate that \method allows for highly efficient sampling of training data for fine-tuning the target downstream task, effectively doubling the value of each real training example.

\begin{figure*}[t]
    \centering
    \includegraphics[width=\linewidth]{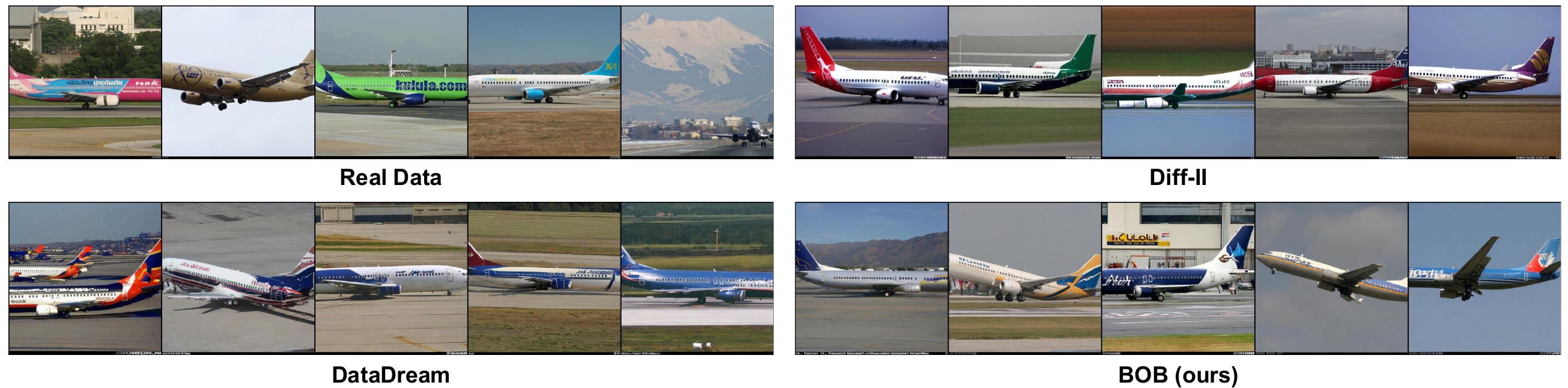}
    \vspace{-15pt}  
    \caption{\textbf{Visualizations.} \textit{left.} 737-400 images from real data and synthetic data generated by Diff-II, DataDream, and \method (ours). Diff-II generates images with aircrafts with high contrast in simple backgrounds. DataDream generates more realistic aircrafts that are only on the ground. Our method \method generate realistic aircrafts in very diverse settings such as taking off, flying, or on the ground with mountainous background, resulting in images that are visually similar to real images. Images for other datasets are shown in the Appendix.}
    \vspace{-15pt}
\label{fig:qualitative}
\end{figure*}

\subsection{Long-tail Classification}
To demonstrate that our method extends beyond few-shot classification, we perform experiments in a long-tail classification setting using the CUB-LT dataset \citep{samuel2021generalized} and Flower-LT \citep{wang2024enhance} using an ImageNet pre-trained ResNet-50 backbone. In the long-tail classification settings, the number of images per class used for fine-tuning is artificially skewed to follow an exponential distribution specified in \citet{samuel2021generalized} for CUB-LT and \citet{wang2024enhance} for Flower-LT. 
For synthetic data generation, we set a budget of 200 total images per class. Similar to the few-shot examples, we duplicate the real examples by a constant factor $c$ such that the number of images in the head (classes with abundant real images) are close to 200 images, arriving at the number of images defined by: $200 - \text{number of real images} \times c$ ($c=6$ for CUB-LT and $c=5$ for Flower-LT). \cref{table:longtail} shows \method outperforms existing methods by a considerable margin on long-tail classes.

\smallsec{Performance gains on CUB-LT.} CUB-LT is a relatively challenging datasets for long-tail classification with relatively lower accuracies when training with only real data. In this challenging setting, we observed across every imbalanced factor on CUB-LT datasets, our method \method outperforms existing ones by a margin of at least 4\%. We observe the greatest improvement in performance when there is a large class imbalance. For the setting with the largest imbalance (IF=100), our method improves from previous method by at least 6\%: 56.10\% with Diff-II to 62.19\% with ours and 53.42\% with DataDream to 63.06\% with ours. Although classes with many examples tend to have similar accuracies (87.54\% with Diff-II vs. 88.48\% with ours), there is a large increase in performance in the classes with few examples: 44.05\% with Diff-II to 52.24\% with ours. These findings provide strong evidence that for challenging long-tail tasks such as CUB-LT, \method generates more informative data than previous methods.

\smallsec{Competitive performance on Flower-LT.} In contrast to the dataset CUB-LT, Flower-LT is a relatively easier task with fewer classes and higher accuracies. For the most difficult part of this benchmark with IF=100, \method achieves a 2\%-3\% gain. For IF=50 and IF=10, \method performs competitively against previous methods with accuracies within 1\% range. These results show that even for easier long-tail tasks, \method  generates the best synthetic datasets.
\vspace{1mm}

\subsection{Analysis}
\label{sec:analysis}
Having demonstrated that our proposed method \method outperforms existing methods for few-shot classification and long-tail classification, we next analyze \emph{why} it works better. We perform analysis using the 10-shot Aircraft data setting with ImageNet-1K pre-trained ResNet-50 backbone to reveal that: (1) our method produces synthetic data that more closely resembles the real data. (2) the performance gains are not due to distilling class knowledge from the captioning model, (3) diversity in the synthetic data is crucial but does not fully explain the performance gains, and (4) both context preservation and marginalization are important for creating high performing data.

\smallsec{Qualitative analysis.} Visualization of images generated by our method shown in \cref{fig:qualitative} produces a sharp contrast compared to existing methods. We observe in \cref{fig:qualitative} that the previous methods lack realism or diversity. With Diff-II, the aircrafts have high contrast with background that is typically monotonous. 
DataDream generates realistic looking images with many complexities in the background but the aircrafts are all on the ground. 
In contrast, our method \method, produces images that are both realistic and diverse; including images of aircraft on the ground with a mountain in the background, an aircraft taking off, etc. It is clear that the synthetic images generated from \method resemble the real images closest: suggesting that the source of performance gain is from better alignment with the real data distribution.

\begin{figure}
    \centering
\includegraphics[width=\linewidth]{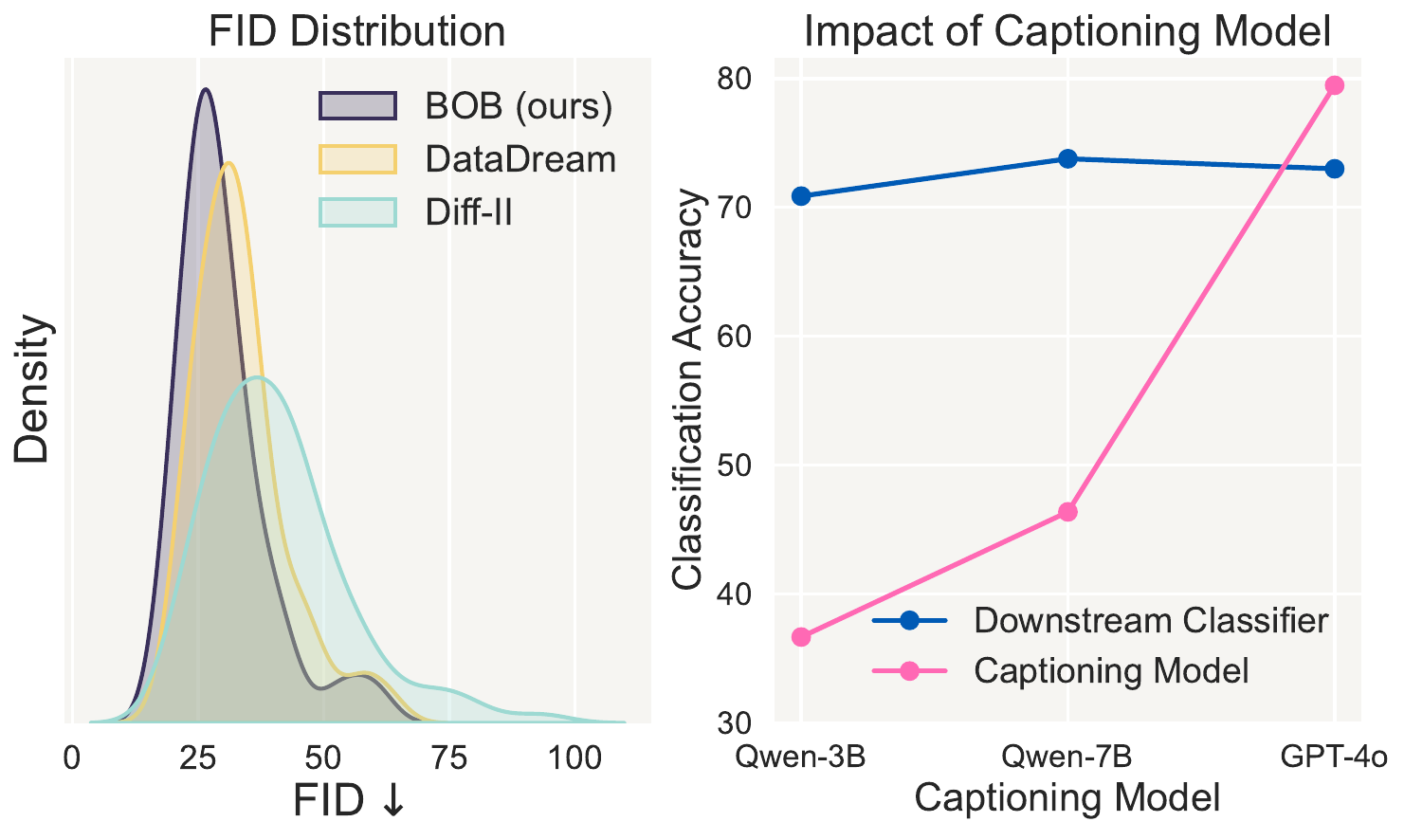}
    \vspace{-20pt}
    \caption{\textbf{Where do the performance gains come from?} \textit{(Left)} Per-class FID density plot shows \method generates data considerably closer to the training distribution than previous methods, explaining the performance gains. \textit{(Right)} Caption model classification accuracy vs. downstream classifier performance shows no correlation, ruling out distillation as the source of gains.}
    \vspace{-15pt}
    \label{fig:density_distillation}
\end{figure}
\vspace{1mm}

\smallsec{Real vs. synthetic distribution.} 
We provide an analysis on the alignment of the synthetic dataset distributions with the target dataset by computing the per-class Frechet Inception Distance (FID) \citep{heusel2017gans} between the whole training dataset and the synthetic datasets. A lower value indicates that the generated dataset is closer to the real data distribution. We reveal that the synthetic dataset produced by \method is better aligned to the training data distribution compared to DataDream and Diff-II. In \cref{fig:density_distillation} \textit{left}, we observe that, on average, the per-class FID is lower for our method \method with the mode at around 26 compared to 31 with DataDream and 37 with Diff-II. 91 out of 100 classes exhibited a decrease in FID using our method \method compared to DataDream (see the Appendix for more details). The lower FID further reinforces our qualitative analysis that the generated data is closer to the real data distribution.

\smallsec{Is it distillation?} Since the captioning models themselves have some fine-grained classification capabilities, the natural question is whether we are inadvertently distilling these capabilities down to the T2I model, and subsequently, to the downstream classification model. To test this hypothesis, we use two additional captioning model: Qwen VL2.5-3B \citep{Qwen2.5-VL} and GPT-4o \citep{hurst2024gpt}. If the source of the performance gains is distillation, then the classification capabilities of these captioning model should be a strong indicator of downstream classification performance. However, \cref{fig:density_distillation} \textit{right} demonstrates that there is no such association at any discernable level. We observe that GPT-4o has considerably higher fine-grained classification capability with accuracy close to 80\%. However, when used as a captioning model in our method, the downstream classification performance does not improve. Similarly for the weaker captioning model Qwen-3B, we do not exhibit a considerable decrease with accuracy 70.88\%, which is still over 3\% higher than the DataDream baseline. These findings suggest that the performance gains we observed are not due to distilling classification capabilities from the captioning model.

\begin{table}[t]
\centering 
\caption{\textbf{Analysis on prompt diversity.} Results show that prompt diversity is very important as GPT generated background and pose performs very well. However, our method \method is still the best, indicating that it isn't all about prompt diversity.}
\vspace{-1em}
\begin{tabular*}{\columnwidth}{@{\extracolsep{\fill}}lc@{}}
\toprule
\textbf{Captions Source} & \textbf{Accuracy} \\ 
\midrule
Background/pose of the same class & 64.38 \\
Suffices provided from Diff-II \citep{wang2025inversion} & 66.53 \\
100 background/pose extracted from GPT & 72.10 \\
All the extracted background/pose (\method) & \textbf{73.78} \\ 
\bottomrule
\end{tabular*}
\vspace{-1em}
\label{table:diversity}
\end{table}

\smallsec{Is it prompt diversity?} Another question is whether the performance gains from context marginalization arise from mitigating unintended inter-class associations or simply from increasing prompt diversity. To disentangle these factors, we introduce three settings. First, we mimic context marginalization but sample backgrounds and poses only from the same class, yielding low diversity while preserving spurious correlations; this leads to very poor performance (64.38\%) in \cref{table:diversity}. Next, we use the Diff-II \citep{wang2025inversion} suffixes, which has 10 dataset-level caption summaries, offering low diversity but remove spurious correlations; performance improves from 64.38\% to 66.53\%, indicating the benefit of breaking these correlations, though the accuracy remains low. Finally, we prompt GPT to generate 100 possible aircraft backgrounds and poses, providing high diversity; this achieves strong performance (72.10\%). While this highlights the importance of diversity, it still underperforms our proposed method, which uses only the contexts present in the small dataset. Overall, prompt diversity is crucial but cannot fully explain the gains of \method.

\smallsec{Ablation studies.} Finally, we ablate on context preservation and marginalization. 
Results from \cref{table:ablation} reveals both preservation and marginalization is necessary for creating high-performing datasets. 
Without preservation and marginalization, the algorithm is identical to DataDream baseline, which achieves only 68\%. 
Adding marginalization without preservation result in a 3\% improvement in accuracy to 70.13 \%. 
Including both marginalization and preservation results in the best accuracy at 73.78\%. Interestingly, preservation without marginalization leads to worse performance. We observe that the performance decreased from 68\% to 65.90\% when using preservation without marginalization. 
These ablations reveal the necessity of both context preservation and dataset-wide marginalization for generating highly informative images. Additional ablations on the attributes used is in the Appendix.

\begin{table}[t]
\centering
\caption{\textbf{Ablations on preservation/marginalization.} Results show both are needed for creating high-performing synthetic data.}
\vspace{-1em}
\begin{tabular*}{\columnwidth}{@{\extracolsep{\fill}}ccc@{}}
\toprule
\textbf{Preservation} & \textbf{Marginalization} & \textbf{Accuracy} \\ 
\midrule
\ding{55} & \ding{55} & 68.00 \\
\checkmark & \ding{55} & 65.90 \\
\ding{55} & \checkmark & 70.13 \\
\checkmark & \checkmark & \textbf{73.78} \\ 
\bottomrule
\end{tabular*}
\vspace{-15pt}
\label{table:ablation}
\end{table}

\section{Conclusion}
\label{sec:conclusion}
We introduce \method as a fine-tuning strategy for text-to-image (T2I) models that mitigates overfitting and addressing the unique challenges of fine-grained classification. By leveraging more detailed captioning to extract class-agnostic background and pose information, conditioning on these features during fine-tuning, and marginalizing them out during data generation, our approach reduces unintended class associations and narrows the distribution gap between synthetic and real data. Extensive experiments across multiple backbones, datasets, and scales demonstrate consistent and significant performance gains, including over 7\% improvement on the Aircraft dataset and best performance in nearly all settings. This work highlights the potential of caption-guided fine-tuning to improve synthetic data for downstream classification tasks and opens avenues for further research on scaling it to broader domains.

\smallsec{Limitation.} \method is designed for fine-grained recognition where class-agnostic attributes can be reliably inferred. The automatic discovery of these attributes or invariances remains an open problem. Additionally, scaling this up to the more complex coarse-grained recognition settings like ImageNet remains challenging: classes become context-dependent, and unintended spurious correlations can simplify the task, thereby inflating task performance.

\section*{Acknowledgements}
This material is based on research supported by the National Science Foundation under Grant No.
2112562 and Solidigm AI SW. Any opinions, findings, and conclusions or recommendations expressed in this material are those of the authors’ and do not necessarily reflect the views of the National Science Foundation.
\section*{Ethics Statement}
\label{sec:ethics}
We use publicly available datasets Aircraft \citep{maji13fine-grained}, CUB \citep{wah2011caltech}, Car \citep{krause20133d}, Pets \citep{parkhi2012cats} as well as CUB-LT \citep{samuel2021generalized} and Flower-LT \citep{wang2024enhance} datasets for non-commercial research purposes, in line with dataset disclaimers (e.g. CUB). These datasets may include biases in representation as the data is limited in terms of their geographical or environmental context, and the models we train with this data are not intended to be used directly without such representation considerations.

The goal of this research is to develop a method for augmenting training for fine-grained classification especially in the few-shot regime. 
Such settings may naturally include biases in the data due to low sample size and the downstream effects of such use must be studied in detail when applying to real-world applications.
In addition, since any such method can be misused to violate privacy and copyrights or creating deepfakes, we encourage use of watermarks for both real images as well as model generated images for their detection and mitigation.

{
    \small
    \bibliographystyle{ieeenat_fullname}
    \bibliography{main}
}

\clearpage
\setcounter{page}{1}
\onecolumn
{
   \newpage
       
        \centering
        \Large
        \textbf{\thetitle}\\
        \vspace{0.5em}Supplementary Material \\
        \vspace{1.0em}
   }

\hypertarget{toc}{}
\startcontents[sections]
\printcontents[sections]{l}{1}{\setcounter{tocdepth}{2}}

\newpage
\section{Additional implementation details}
\label{sec:add_implement}
\begin{table}[t]
\centering
\caption{\textbf{Hyperparameters for fine-tuning.} Hyperparameters used for fine-tuning of T2I model and downstream classifier. A list of parameters indicates the hyperparameter sweep using the validation set. The number of epochs indicated in \textit{parenthesis} is the epochs used for fine-tuning on test set with the best hyperparameter on validation set.}
\renewcommand{\arraystretch}{1.3} 
\resizebox{0.85\textwidth}{!}{%
\begin{tabular}{lcccc}
\hline
                & T2I       & CLIP                         & ImageNet               & MAE              \\ \hline
Learning rate   & 1e-4      & {[}1e-4, 1e-5, 1e-6, 1e-7{]} & {[}1e-3, 1e-4, 1e-5{]} & {[}1e-3, 5e-4{]} \\
Weight decay    & 1e-2      & {[}5e-4, 1e-4{]}             & {[}0.01, 1e-4, 0{]}    & 0.05             \\
Layer decay     & -         & -                            & -                      & {[}0.65, 0.75{]} \\
$\lambda$          & -         & {[}0.5, 0.8{]}              & {[}0.5, 0.8{]}       & {[}0.5, 0.8{]}   \\
Epochs          & 400       & 10 \textit{(100)}       & 10 \textit{(100)}  & 10 \textit{(100)}         \\
Batch size      & 80        & 64                           & 64                     & 64               \\
Scheduler       & Cosine    & Cosine                       & Cosine                 & Cosine           \\
Warm up         & 100 steps & 3 epochs                     & 3 epochs               & 5 epochs         \\
Max norm        & 1.0       & -                            & -                      & -                \\
LoRA rank       & 16        & 16                           & -                      & -                \\
Mixed precision & No        & fp16                         & fp16                   & fp16             \\ \hline
\end{tabular}
}
\label{table:hyperparameters}
\end{table}
\subsection{Hyperparameter sweep} In this section, we go over the hyperparameter used to produce the results in \cref{table:fewshot} and \cref{table:longtail}. The full hyperparameters are listed in \cref{table:hyperparameters}. 
For fine-tuning the T2I model on the few-shot or long-tail images, we follow the procedure in DataDream paper \citep{kim2024datadream} with two difference: using dense captions for text input following our template outline in \cref{sec:context_preservation}, increasing the number of epochs from 200 to 400. 
For CLIP fine-tuning, we follow the pipeline in DataDream paper \citep{kim2024datadream}  
We optimize LoRA layers in both the image and text encoder with rank 16. 
We sweep over learning rate \{1e-4, 1e-5, 1e-6, 1e-7\} and weight decay \{5e-4, 1e-7\}. 
The only difference is that we have an additional sweep for $\lambda$ \{0.5, 0.8\}. 
This results in 16 different configurations we are sweeping over for CLIP fine-tuning. 
For ImageNet fine-tuning, we sweep over learning rate \{1e-3, 1e-4, 1e-5\}, weight decay \{0, 0.01, 1e-4\}, and $\lambda$ \{0.5, 0.8\}.
This results in 18 different configurations for ImageNet. 
For MAE fine-tuning, we use the fine-tuning recipe provided by the original authors. 
We sweep over base learning rate \{1e-3, 5e-4\} and layer-wise learning rate decay \{0.65, 0.75\} due to a discrepancy in the values provided by the original paper and default in their Github release. 
We also sweep over $\lambda$ \{0.5, 0.8\}. In all, this results in 8 different configurations for MAE fine-tuning.

We fine-tune the model for 10 epochs and select the configuration that results in the best validation accuracy across the 10 epochs. 
Using the hyperparameters that gave the best accuracy on the validation set, we fine-tune the original pre-trained model again for 100 epochs. We report the test accuracy for the epoch that corresponds to the best validation accuracy during this training.

\subsection{Parameters for dataset generation}
\begin{wraptable}{r}{0.3\textwidth}
    \vspace{-3em}
    \centering
    \caption{Generation parameters.}
    \renewcommand{\arraystretch}{1.3}
    \setlength{\tabcolsep}{15pt}
    \resizebox{\linewidth}{!}{%
        \begin{tabular}{lc}
\hline
Hyperparameter  & Value \\ \hline
Guidance scale  & 2.0   \\
Number of steps & 50    \\
Mixed precision & fp16  \\ \hline
\end{tabular}
    }
    \label{table:generation}
\end{wraptable}

We follow the same parameters used in DataDream for generating the synthetic dataset shown in ~\cref{table:generation}: guidance scale of 2.0, 50 inference steps, and fp16 mixed precision. The scheduler used is the default for Stable Diffusion v1.5 and Stable Diffusion v2.1-base. 

\smallsec{Other methods.} For generating synthetic data for baselines used for comparisons, we use the default parameters used for data generation provided by their paper. For the Diff-Mix method, an additional CLIP filtering is used to remove problematic images as outlined in the original paper.

\subsection{Creation of validation dataset} 
For fair comparison, we perform hyperparameter tuning on the learning rate, weight decay, and the $\lambda$ hyperparameters when evaluating downstream classification performance. In order to achieve this, we created our own validation sets. In the few-shot classification setting, since not all of the training data is used, we randomly select 16 non-overlapping images per class for FGVC-Aircraft and Oxford-Pets, 10 random non-overlapping images for Stanford-Cars and CUB. However, in the long-tail setting (CUB-LT and Flower-LT), there exist some classes where most of the examples are used for training. Therefore, it is no longer possible to create a separate validation set. Instead, we split the test set into a smaller test set and a held-out validation set with five images per class.

\section{Additional analysis}
\subsection{Number of training epochs}
\label {section:epochs}

\begin{wrapfigure}{r}{0.25\textwidth}
    \vspace{-5mm}
    \centering
\includegraphics[width=\linewidth]{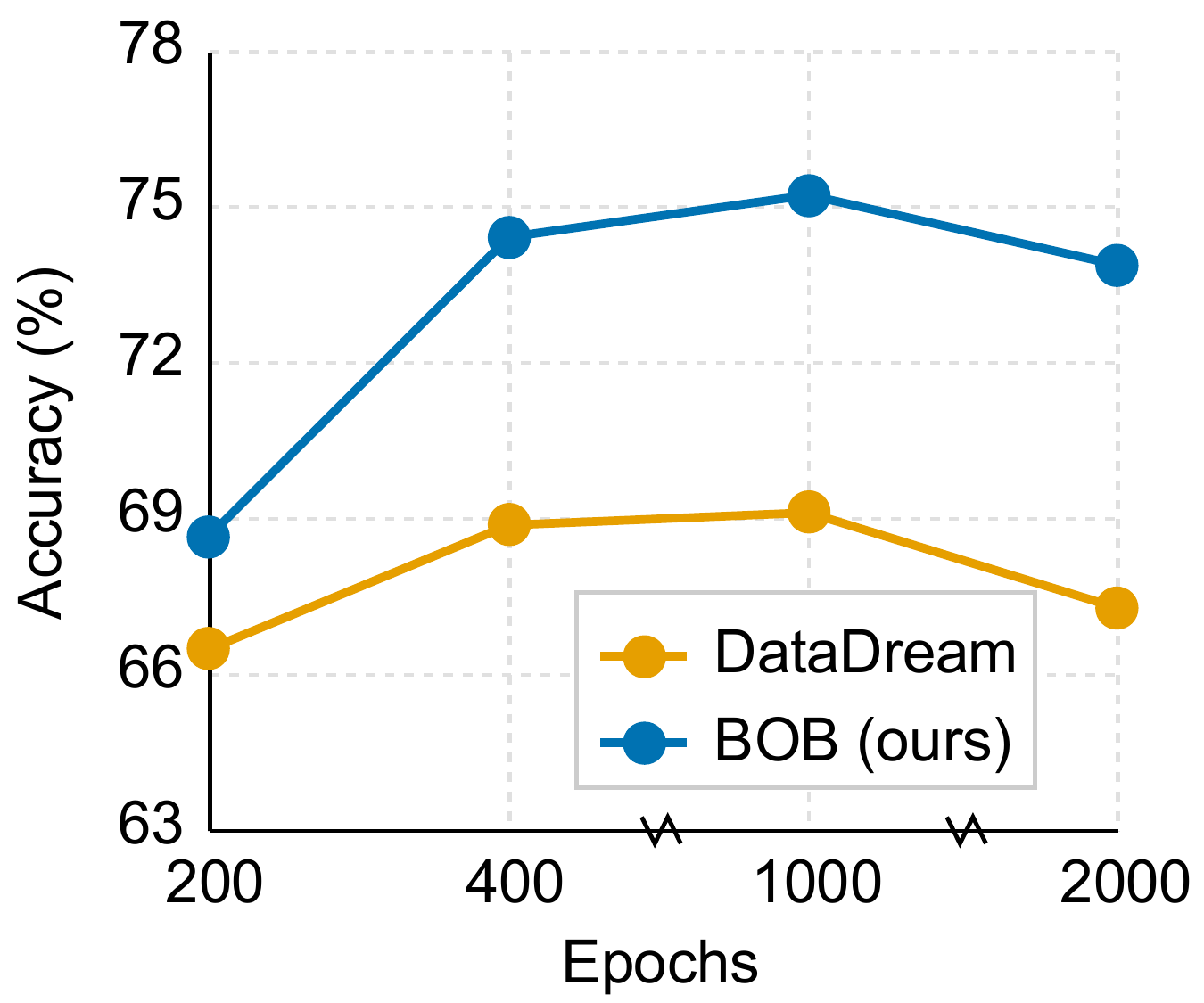}
    \vspace{-5mm}
    \caption{Fine-tuning the T2I model longer helps.}
    \vspace{-2mm}
    \label{fig:epochs}
\end{wrapfigure}
In comparison to DataDream, we increase the number of epochs from 200 to 400. 
To motivate this design decision, we fine-tune a model for 2,000 epochs on 10-shot Aircraft and save the intermediate checkpoints to study the effect of longer fine-tuning towards the generation of informative samples. 
For the intermediate checks, we follow the same synthetic data generation and hyperparameter tuning procedure to obtain the final test accuracy. 
The result shown in \cref{fig:epochs} shows similar effect from number of epochs on both our method \method and DataDream. 
Going from 200 epochs to 400 epochs, DataDream performance improves by 2\%, from 66.49\% to 68.87\%. 
However, our method exhibits a \emph{6\% increase} from 68.65\% to 74.41\%. 
The considerably larger increases suggests that, while DataDream benefits from longer fine-tuning, our method \method benefits from it more. 
Similarly for both methods, the performance peaks at the checkpoint fine-tuned for 1,000 epochs before it starts to decrease again using the 2,000 epoch checkpoint. 
Finally, at every epoch in \cref{fig:epochs}, our method \method outperforms DataDream. 

\subsection{Number of synthetic images}
\label{sec:num_synthetic}
\begin{wrapfigure}{l}{0.25\textwidth}
    \vspace{-2mm}
    \centering
\includegraphics[width=\linewidth]{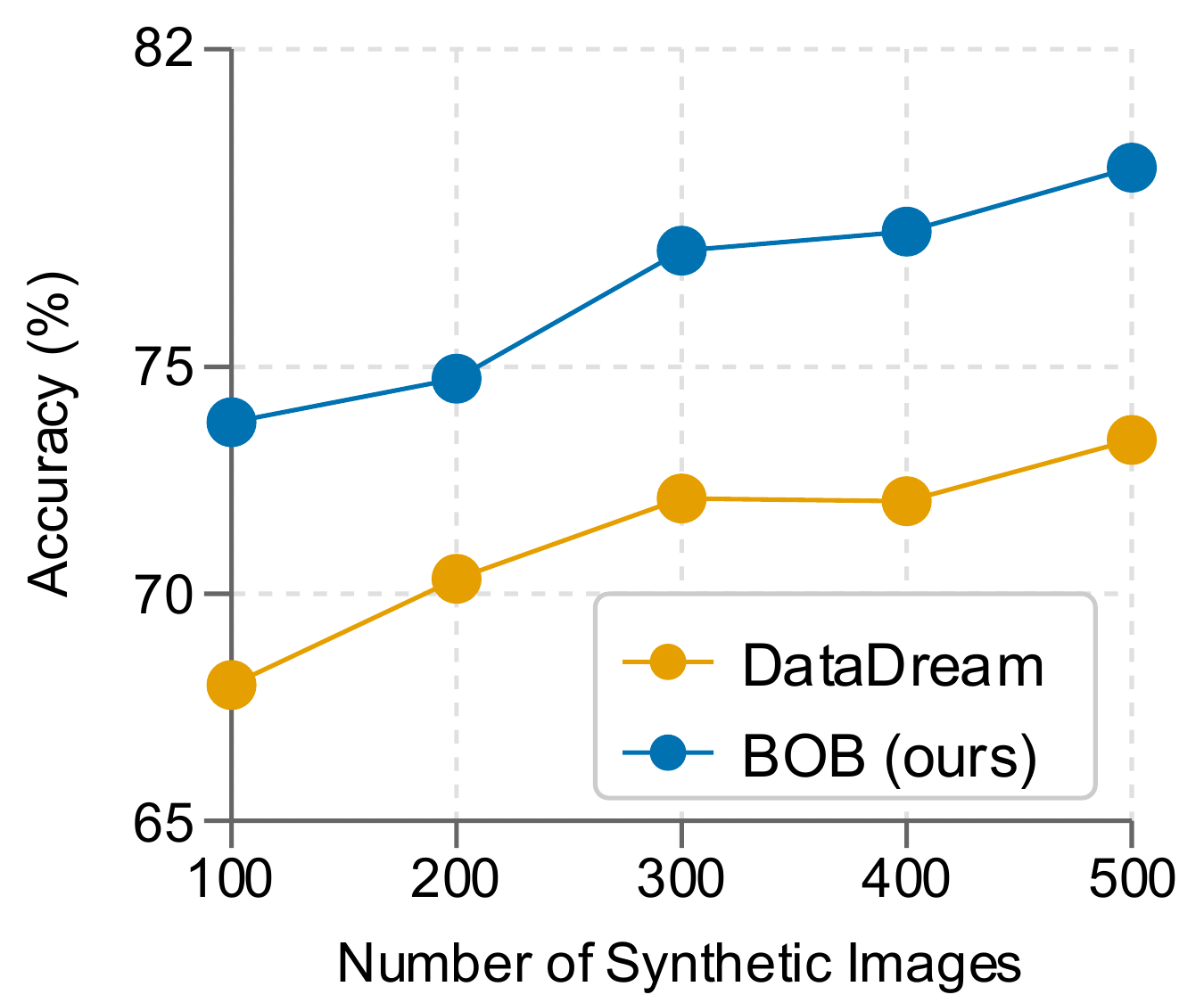}
    \vspace{-5mm}
    \caption{Ablations of number of synthetic images.}
    \label{fig:num_syn}
\end{wrapfigure}
All of the experiments so far are evaluated using 100 synthetic images per class to reduce the generation time and storage requirement. However, we are interested in understanding the effect of increasing the number of synthetic images on our proposed method. Since DataDream is the most performant baseline, we will compare our method \method with DataDream in this ablation. For this ablation, we use the 10-shot Aircraft setting with the ImageNet Pre-trained ResNet-50 backbone and increase the number of images generated per class from 100 to 200-500. We perform the same procedure where hyperparameter sweep on the validation set and final evaluation on the test set. However, to maintain the same amount of computation time, i.e. number of gradient updates, we decrease the number of epochs from 100 to 50, 33, 25, 20 accordingly. This is done to reduce the compute necessary and remove the confounding factors of extra gradient updates. The result shown in \cref{fig:num_syn} reveals that adding more synthetic images increases overall performance and our method \method outperforms DataDream regardless of the number of synthetic images.

\subsection{Number of real images}
\label{sec:num_real}
\begin{wrapfigure}{r}{0.25\textwidth}
    \vspace{-5mm}
    \centering
\includegraphics[width=\linewidth]{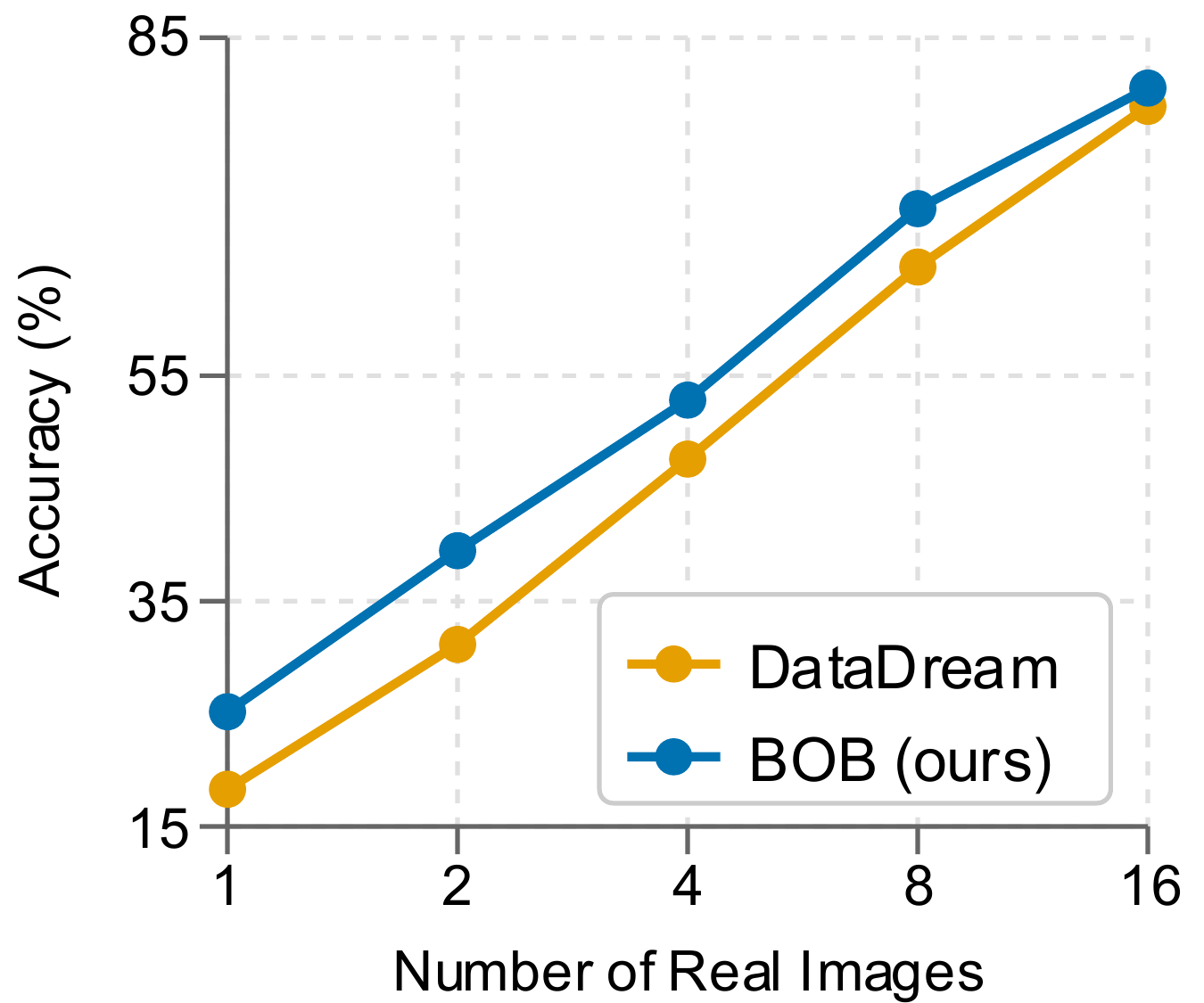}
    \vspace{-5mm}
    \caption{Ablations of number of synthetic images.}
    \vspace{-2mm}
    \label{fig:num_real}
\end{wrapfigure}
In addition to increasing the number of synthetic images, we would like to understand how well our method performs as the number of available real-images increase. To study this, we rerun the whole pipeline on the Aircraft setting using 1, 2, 4, 8,and 16 real images during T2I fine-tuning and classification training. Similar to the original setup, we generate 100 synthetic images per class. The results shown in \cref{fig:num_real} reveals that our method \method outperforms Datadream across all the settings but the performance gap does decrease as we increase the number of real images. In particular, we see that with only 1 image per class, \method achieves an accuracy of 25.18\% while Datadream achieves an accuracy of 18.3\%. In contrast, when we have 16 images per class, \method achieves 80.53\% while datadream achieves 78.91\%. These findings reveals the considerable advantage our method \method presents in very data-constrainted settings. 

\subsection{Attributes ablations}
\begin{wraptable}{r}{0.4\textwidth} 
\centering
\vspace{-1.5em} 
\caption{Ablations on the attributes.}
\vspace{-1em}
\setlength{\tabcolsep}{2pt} 
\small
\begin{tabular}{lcc|cccc}
\toprule
 & \multicolumn{2}{c|}{\textit{Sampling}} & \multicolumn{4}{c}{\textit{Attributes}} \\
 & cpl. & ind. & pose & bg & b+p & +col \\
\midrule
Acc. (\%) & 60.31 & 60.16 & 58.50 & 58.73 & 60.31 & 61.09 \\
\bottomrule
\end{tabular}
\label{table:additional_ablations}
\end{wraptable}
We perform additional ablations studies on the attributes used for marginalization on the 5-shot Aircraft setting with the ImageNet-pretrained backone. First, we study whether marginalizing out background and pose jointly vs. independently makes any difference. We ran our method with pose and background independently sampled and the results shown in \cref{table:additional_ablations} (\textit{Sampling experiments.}) reveals similar performance (coupled: 60.31\% vs. independent: 60.16\%). Next, we study the effect of adding a color attribute, an example attribute that is applicable to aircraft but not to CUB. Results incorporating color or excluding background/pose shown in \cref{table:additional_ablations} (\textit{Attributes experiments.}). We observe that the performance progressively increases as we add more attributes.

\subsection{Stability across different seeds} 
\begin{wraptable}{r}{0.4\textwidth}
\centering
\caption{Consistent claims across different seeds}
\vspace{-1em}
\begin{tabular}{llll}
\hline
           & Seed1          & Seed2          & Seed3 \\ \hline
DataDream  & 52.19          & 54.35          &     55.33  \\
BOB (ours) & \textbf{57.15} & \textbf{59.56} &   \textbf{60.55}    \\ \hline
\end{tabular}

\label{tab:seeds}
\end{wraptable}
To demonstrate that the claims made in the main paper is consistent across different runs, we rerun the 5-shot aircraft settings on ImageNet backbone three additional times (completely different few-shot images) and compare our method BOB with the previous SOTA Datadream. Result in \cref{tab:seeds} reveals consistent findings across three seeds where our method BOB outperforms previous state-of-the-art method DataDream, strengthening our conclusion. 

\subsection{Class differences in real vs. synthetic distribution}
\label{sec:classwise_fid}
\begin{wrapfigure}{r}{0.4\textwidth}
    \vspace{-5mm}
    \centering
\includegraphics[width=\linewidth]{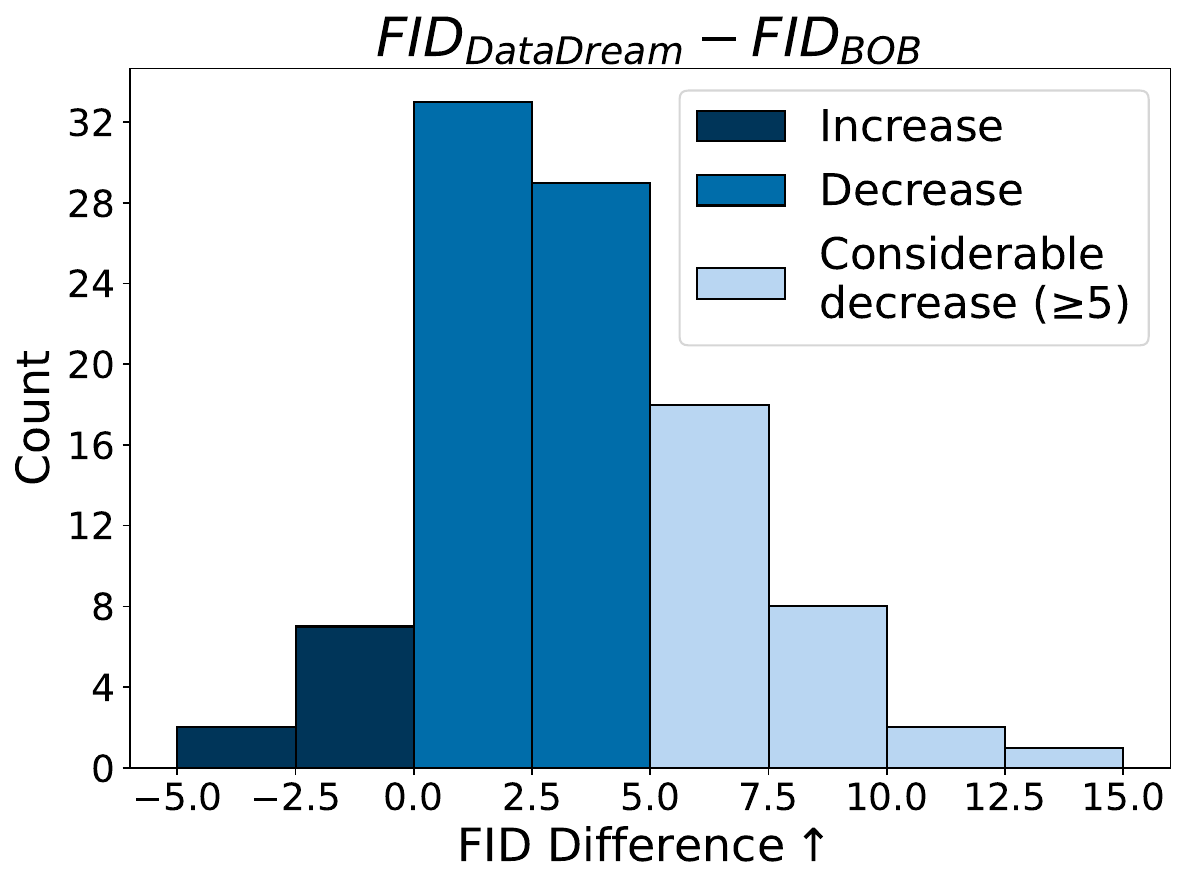}
    \vspace{-5mm}
    \caption{Histogram of the FID difference from DataDream vs. \method for each class.}
    \vspace{-5mm}
    \label{fig:fid_difference}
\end{wrapfigure}
Following our analysis in Section \ref{sec:analysis} towards comparing the distribution of sythetic dataset against the real dataset, we directly compare the FID of each class between synthetic data generated by DataDream and our method \method. We plot the histogram in \cref{fig:fid_difference}. We observe that there are for 91 classes out of 100 FID is lower for our method compared to DataDream. 
Of these, for 29 classes, FID decreases by over 5. 
For all of the classes where our method had a higher FID, the increase is less than 5. 
This means that there are 3x as many classes that observed a considerable decrease in FID than the classes with a relatively low FID increase. The highest increase in FID due to our method is 3.44, which is notably lower than 5. 
This result suggests that our method provides an improvement in FID across all the classes: FID either a considerable increase or remain similar (within 5 FID). 

\subsection{Qualitative examples from Car, CUB, and Pets}
\begin{figure*}[b]
    \centering
    \includegraphics[width=\linewidth]{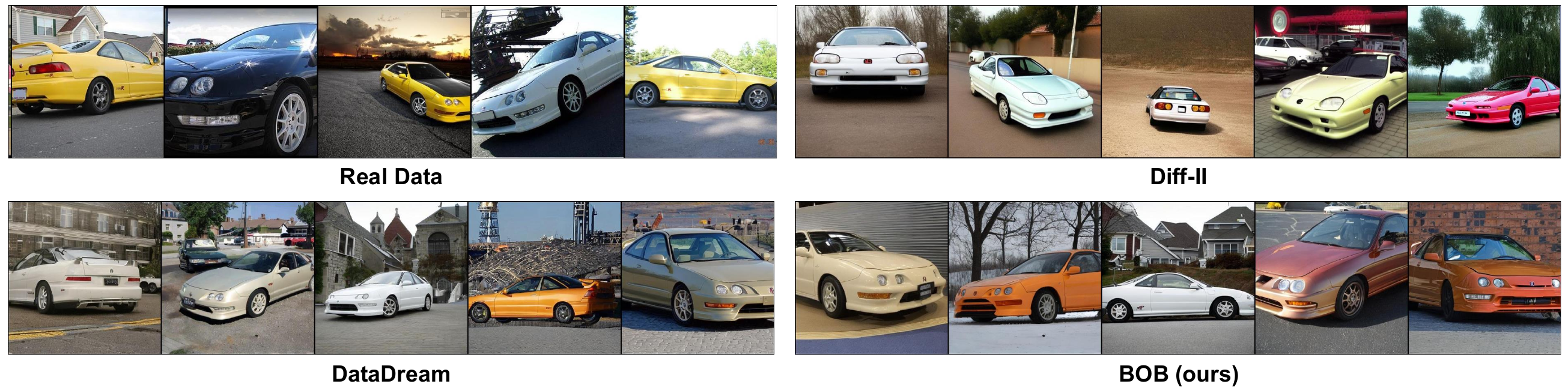}
    \vspace{-15pt}  
    \caption{\textbf{Visualizations on the Car dataset.} Images of the Acura Integra Type R 2001 across different data sources.}
    \vspace{-15pt}
\label{fig:qualitative_car}
\end{figure*}

\begin{figure*}[t]
    \centering
    \includegraphics[width=\linewidth]{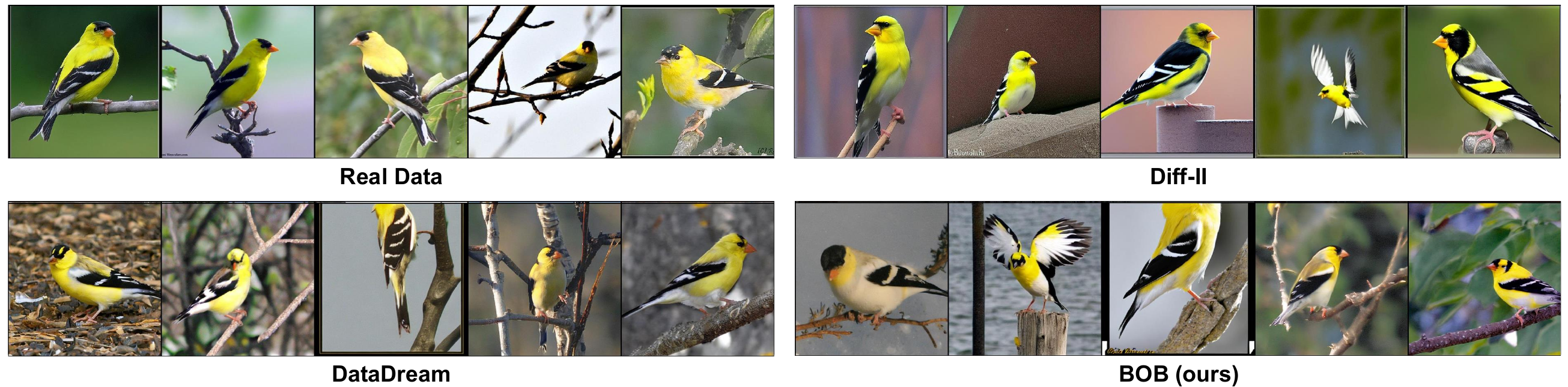}
    \vspace{-15pt}  
    \caption{\textbf{Visualizations on the CUB dataset.} Images of the American Goldfinch across different sources. }
\label{fig:qualitative_cub}
\end{figure*}

\begin{figure*}[t]
    \centering
    \includegraphics[width=\linewidth]{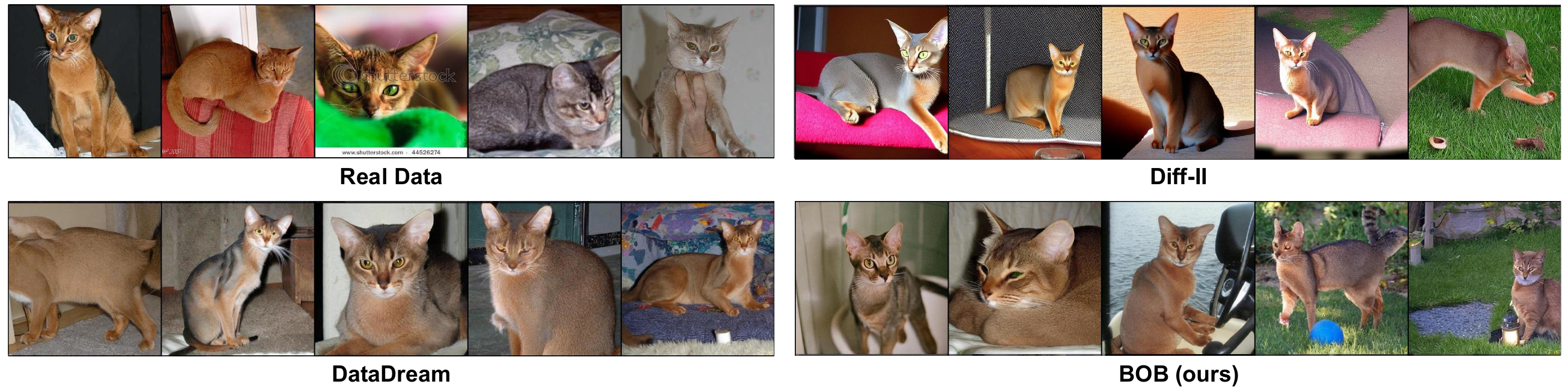}
    \vspace{-15pt}  
    \caption{\textbf{Visualizations on the Pets dataset.} Images of the Abyssinian across different sources.}
\label{fig:qualitative_pets}
\end{figure*}

We include more qualitative examples from the Car, CUB, and Pets dataset shown in \cref{fig:qualitative_car}, \cref{fig:qualitative_cub}, \cref{fig:qualitative_pets}. The examples reveals a similar trend to our qualitative examples of the Aircraft dataset shown in \cref{fig:qualitative}, where our method \method produces more diverse and realistic synthetic images.

\subsection{Significant overlaps in the Pets dataset}
\label{sec:pets}
Recall from \cref{table:fewshot} that on the Pets dataset, fine-tuning a classifier on synthetic data generated images results in very little performance gains across all of the methods in the CLIP and ImageNet classification pre-trained ResNet-50 backbone. 
In the case of CLIP, there is \emph{no performance gains} with all of the methods arriving at an accuracy within 1\% compared to just fine-tuning on real data. 
This is because the zero-shot classification accuracy on Pets using CLIP is already 91\% as reported in the DataDream paper \citep{kim2024datadream}. As a result, the model already has very strong classification capabilities, and therefore, additional synthetic data isn't as impactful, if at all. For the ImageNet classification trained backbone, we make a similar observation where most of the 37 pet classes are already present in ImageNet. To study this, we manually search up the pet names (as well as adjacent names since same pets have multiple names) in the ImageNet classes. The result is shown in \cref{table:pets_imagenet}. We observe that 25 of the 37 classes have a corresponding ImageNet class. Similar to the CLIP setting, if the backbone very high classification capabilities, then it is not a good evaluation metric for determining the strength of classification signals in the synthetic dataset. These findings explains the different trends observed in the MAE setting vs. ImageNet or CLIP setting from the Pets dataset in \cref{table:fewshot}. 
\begin{longtable}{ll ll}
\caption{\centering Oxford-IIIT Pets classes with ImageNet IDs (– if not present) \label{table:pets_imagenet} } \\
\toprule
\textbf{Pet name} & \textbf{ImageNet ID} & \textbf{Pet name} & \textbf{ImageNet ID} \\
\midrule
\endfirsthead

\toprule
\textbf{Pet name} & \textbf{ImageNet ID} & \textbf{Pet name} & \textbf{ImageNet ID} \\
\midrule
\endhead

\bottomrule
\endfoot

Abyssinian & – & American Bulldog & - \\
\rowcolor{gray!10} American Pit Bull Terrier & n02093428 & Basset Hound & n02088238 \\
Beagle & n02088364 & Bengal & – \\
\rowcolor{gray!10} Birman & – & Bombay & – \\
Boxer & n02108089 & British Shorthair & – \\
\rowcolor{gray!10} Chihuahua & n02085620 & Egyptian Mau & n02124075 \\
English Cocker Spaniel & n02102318 & English Setter & n02100735 \\
\rowcolor{gray!10} German Shorthaired & n02100236 & Great Pyrenees & n02111500 \\
Havanese & – & Japanese Chin & n02085782 \\
\rowcolor{gray!10} Keeshond & n02112350 & Leonberger & n02111129 \\
Maine Coon & – & Miniature Pinscher & n02107312 \\
\rowcolor{gray!10} New Found Land & n02111277 & Persian & n02123394 \\
Pomeranian & n02112018 & Pug & n02110958 \\
\rowcolor{gray!10} Ragdoll & – & Russian Blue & – \\
Saint Bernard & n02109525 & Samoyed & n02111889 \\
\rowcolor{gray!10} Scottish Terrier & n02097298 & Shiba Inu & - \\
Siamese & n02123597 & Sphynx & – \\
\rowcolor{gray!10} Staffordshire Bull Terrier & n02093256 & Wheaten Terrier & n02098105 \\
Yorkshire terrier & n02094433 & & \\

\end{longtable}

\end{document}